\documentclass[conference]{IEEEtran}

\usepackage{times}
\usepackage{epsfig}
\usepackage{graphicx}
\usepackage{amsmath}
\usepackage{amssymb}
\usepackage{multirow}
\usepackage{setspace}
\usepackage{subfigure}
\usepackage{balance}
\usepackage{url}

\usepackage{array}
\newcolumntype{V}{>{$\vcenter\bgroup\hbox\bgroup}c<{\egroup\egroup$}}

\usepackage{amsmath,amstext,amsfonts,amsxtra,amssymb,latexsym,bm,color}
\usepackage{psfrag,balance}
\usepackage{array}
\usepackage{multirow}
\usepackage{url}

\newcommand{\diag}{{\mathsf{diag}}}

\newcommand{\bs}[1]{{\mathbf{#1}}}

\graphicspath{{./imgs/}}

\begin{document}

\title{Achieving Stable Subspace Clustering\\by Post-Processing Generic Clustering Results}

\author{\begin{tabular}{ccc}
          Duc-Son Pham & Ognjen Arandjelovi\'c & Svetha Venkatesh\\
          Department of Computing & School of Computer Science & School of Information Technology\\
          Curtin University & University of St Andrews & Deakin University\\
          Australia & United Kingdom & Australia\\
          & & \\
          & & \\
        \end{tabular}}
        
\maketitle

\begin{abstract}
We propose an effective subspace selection scheme as a post-processing step to improve results obtained by sparse subspace clustering (SSC). Our method starts by the computation of stable subspaces using a novel random sampling scheme. Thus constructed preliminary subspaces are used to identify the initially incorrectly clustered data points and then to reassign them to more suitable clusters based on their goodness-of-fit to the preliminary model. To improve the robustness of the algorithm, we use a dominant nearest subspace classification scheme that controls the level of sensitivity against reassignment. We demonstrate that our algorithm is convergent and superior to the direct application of a generic alternative such as principal component analysis. On several popular datasets for motion segmentation and face clustering pervasively used in the sparse subspace clustering literature the proposed method is shown to reduce greatly the incidence of clustering errors while introducing negligible disturbance to the data points already correctly clustered. 
\end{abstract}

\IEEEpeerreviewmaketitle

\section{Introduction}
\label{SEC_INTRO}

Subspace clustering is an important unsupervised learning research topic. By modelling the distribution of data as a union of subspaces \cite{ElhaVida2013,AranCipo2006e} multiple subspace models improve on the single subspace assumption \cite{CandLiMaWrig2011,CandRech2009,Aran2014} and more meaningfully capture the physical structure of the underlying problem. For example, texture features of different image regions are represented well by a mixture of Gaussian distributions \cite{YangWrigMaSast2008} in natural image segmentation. In motion segmentation tracked points of different rigid body segments are naturally divided into subspaces whose dimensions are bounded from above by a fixed number resulting from the corresponding motion equations \cite{CostKana1998,VidaTronHart2008}. Frontal face images have also been shown to lie in subject-specific 9-dimensional subspaces under the Lambertian reflectance model \cite{ElhaVida2013,LeeHoYangKrie2005}. Multiple subspace modelling is relevant to numerous data mining problems as well, such as gene expression \cite{KrieKrogZime2009}. 

A variety of different approaches to subspace clustering have been described in the literature; the reader is referred to \cite{ElhaVida2013} for a comprehensive review. However recent work has mainly centred around using the ideas of sparsity and low rank representations, which have been major algorithmic successes in compressed sensing \cite{CandRombTao2006,Dono2006} and matrix completion \cite{CandLiMaWrig2011,CandRech2009}.The first work in this direction is sparse subspace clustering (SSC) \cite{ElhaVida2009} which regularizes the model-fitting term with the $\ell_1$ norm on the self-expressiveness coefficients, thereby promoting sparsity. There are several advantages of this approach. Firstly it alleviates the need for the knowledge of the number of subspaces and their dimensions in advance, which was required by previous approaches. Secondly the convex formulation can be extended or tailored to specific needs, e.g.\ to handle corrupted or missing data. Finally its convex formulation can be solved with a practically satisfactory accuracy using a framework known as alternating directions method of multipliers (ADMM) \cite{BoydPariChuPele+2011}. Another related alternative is trace norm regularization \cite{LiuLinYu2010} which seeks sparsity in the transform domain instead. This subspace clustering approach has attracted a large amount of theoretical analysis \cite{NasiHart2011,SoltCand2012}, work on performance improvement with spatial constraints \cite{PhamBudhPhun+2012}, combined low-rank and sparsity regularization \cite{WangXuLeng2013}, group sparsity regularization \cite{SahaPhamPhunVenk2013}, scalability \cite{PengZhanYi2013}, thresholding ridge regression \cite{PengYiTang2015}, multi-view input \cite{CaoZhanFuLiu+2015}, mixtures of Gaussians \cite{LiZhanLinLu2015}, or latent structure \cite{LiVida2015}. 

Though many extensions of sparse subspace clustering have demonstrated promising results in different applications, and theoretical analysis has established useful results to explain its success, a number of practical challenges remain. As with any unsupervised method, selecting the right parameters to obtain an optimal performance is not a trivial task. The default values for parameters in most of publicly available code implementing SSC variants are unlikely to give optimal clustering results in terms of either accuracy or normalized mutual information for every dataset. Even for a specific dataset, the optimal parameter values also depend on the number of classes, because the ideal self-expressiveness coefficients have a fraction of non-zero entries inversely proportional to the number of clusters. The reported results in much of the previous on sparse subspace clustering are for optimal settings which may be difficult to obtain in practice. 

In this paper we propose a novel technique to improve the performance achieved by conventional sparse subspace clustering approaches. It can be used as a post-processing step to re-assign samples to more suitable clusters. It can also be seen as analogous to cross validation in supervised learning. Our idea is to re-examine the subspace assumption: if a data sample truly belongs to the subspace induced by the points in the corresponding cluster, it must be distant from other clusters. If this requirement is not met, the data point is better re-assigned to another more suitable cluster. Our key technical contribution lies in a novel algorithm that computes regularized and stable subspaces from data points of initial clusters. To do so, we use an idea from a powerful framework in statistics known as stability selection. Using the computed projection onto the regularized and stable subspaces, we re-examine the $\ell_p$ norms of residual vectors of all data points and re-assign them to suitable subspaces accordingly. We demonstrate the usefulness of the proposed method in improving SSC and many variants on popular face clustering and motion segmentation data sets under different scenarios. Our results and the analysis thereof suggest that the proposed method is a highly useful and non-application specific post-processing technique which can be applied following any subspace clustering algorithm.

The paper is organized as follows. In Section~II, we review related works on robust subspace estimation. Section~III details the proposed method. Section~IV studies how the proposed method improves preliminary results obtained by popular subspace clustering algorithms. Finally, Section~V concludes the paper. 

\section{Related Work}
There are a number of different approaches to robust model fitting in the literature that are potentially useful for the problem considered in this work.  One of the most frequently used methods for robust feature matching in computer vision is RANSAC (RANdom SAmple Consensus) \cite{FiscBoll1981}. Within this general framework, a generative model must be specified \emph{a priori} as well as the minimum number of samples as a threshold for goodness of fit (inlier count). The algorithm iterates between obtaining a model parameter estimate based on a random subset of data points and counting the number of inliers based on the obtained model over all data points, until that count exceeds the specified threshold. The main advantages of RANSAC are its simplicity, wide applicability to many problems, and ability to cope with a large (up to 50\%) portion of outliers. However, model fitting does not generally improve with iterations and the algorithm may need a substantial number of iterations to find the optimal model. In addition, it can only estimate one model at a time.

Another random sampling based approach frequently used in statistics and signal processing, which might be useful for determining the incorrectly assigned data points in a cluster, is the bootstrap. An inference about a given set of data points is produced by computing the empirical distribution of the bootstrap samples, which are generated from a model constructed from the data itself. Thus the algorithm can determine which data points most likely deviate from a given model. In computer vision, the bootstrap is often used in building a background model in order to detect foreground objects \cite{HughGrzeGree2013}. Its variant in machine learning is bagging \cite{Brei1996}, which was originally derived for decision trees. The method most closely related to that proposed herein is bagging PCA \cite{LengChenYuanBai+2014}. However, it is different to the current work in two major aspects: the sampling mechanism (bagging vs.\ random sampling without replacement), and the consolidation of the final results (taking the union rather than average). Whilst bagging is also possible, we found that it does not provide any advantage in performance compared to our approach, and that it needs to collect many more resamples. This necessitates an increase in computational cost. 

Stability selection \cite{MeinBuhl2010} is another recent statistical method, primarily designed for variable selection in regression problems. The core idea of stability selection is to accumulate selection statistics over random subsets of the original data. This allows the experimenter to decide which response variables are most relevant to the regression problem. Stability selection is suitable for high-dimensional data where estimation of structure is difficult due to the dimensionality of the variables. 

Another approach to extracting a robust subspace from a set of data samples in the presence of outliers is matrix completion \cite{CandLiMaWrig2011,CandRech2009} which uses convex optimization to extract the underlying data structure. Here, the data matrix is expressed as a sum of low-rank and corruption parts. The low-rank component models the intrinsic subspace where the data lies in, and the corruption term captures the deviation from that subspace assumption. For a suitable model of the corruption, it is possible to parameterize outliers explicitly \cite{XuCaraSang2012}. However, this method is limited to one subspace at a time. Besides, it is hard to select an optimal parameter without prior knowledge and the method can perform poorly when the preliminary subspace clustering is not sufficiently accurate. Thus, it is generally of limited use as a technique for the post-processing of preliminary clustering results.

\section{Proposed Method}\label{SEC_METHOD}
Our method learns stable subspaces from a preliminary clustering, and then uses computed projections onto these subspaces to improve the result by reassigning some of the data points to more suitable clusters. Like in sparse subspace clustering or indeed any other alternative, extracting a stable subspace is the key to success. We approach the challenge as an anomaly detection problem \cite{AranPhamVenk2015,Aran2011a}. Inspired by subspace analysis based methods for anomaly detection \cite{JackMudh1979,Jack1959,PhamVenkLazaBudh2014}, we propose to use the principal subspace of each cluster as a measure of the span of its data points. Thus, the fitness of a data point to each cluster is quantified by its projection on the residual subspace.  We also use the random sampling mechanism to obtain a more stable principal subspace. 

Let us denote as $\bs{X}=[\bs{x}_1,\ldots,\bs{x}_n]$ the original data matrix where for each data point it holds that $\bs{x}\in\mathbb{R}^d$. Suppose that there are $K$ clusters, and denote as $\mathcal{S}^0_k, k=1,\ldots,K$ the ground-truth index sets of these clusters. The corresponding data sub-matrix of cluster $k$ is denoted as $\bs{X}_{\mathcal{S}^0_k}$ which is defined as a collection of data points $\bs{x}_i, i\in\mathcal{S}^0_k$. Similarly, denote as $\mathcal{S}_k, k=1,\ldots,K$ the index sets from preliminary clustering obtained by SSC; thus the data points of the $k$-th cluster are $\bs{X}_{\mathcal{S}_k}$. The true subspace associated with cluster $k$ is therefore $\mathsf{span}(\bs{X}_{\mathcal{S}^0_k})$ and the $k$-th subspace estimated by SSC is $\mathsf{span}(\bs{X}_{\mathcal{S}^0_k})$. Here, $\mathsf{span}$ denotes the span of a set of vectors, i.e.\ the subspace created by all possible linear combination of vectors in that set.

When the preliminary clustering is not perfect, it is expected that $\mathcal{S}^0_k \setminus  {\mathcal{S}^0_k \cap \mathcal{S}_k} \neq \emptyset$. So we can write:
	\begin{eqnarray}
	 	\bs{X}_{\mathcal{S}^0_k} & = & \bs{X}_{\mathcal{S}^0_k \cap \mathcal{S}_k} \cup \bs{X}_{\mathcal{S}^0_k \setminus  ({\mathcal{S}^0_k \cap \mathcal{S}_k})} ,\\
		\bs{X}_{\mathcal{S}_k} & = & \bs{X}_{\mathcal{S}^0_k \cap \mathcal{S}_k} \cup \bs{X}_{\mathcal{S}_k \setminus  ({\mathcal{S}^0_k \cap \mathcal{S}_k})} .
	\end{eqnarray}
Here, $\mathcal{S}^0_k \cap \mathcal{S}_k$ represents the indices of the correctly clustered data points. Our idea is that instead of estimating the full subspace, we only approximate the subspace spanned by $\bs{X}_{\mathcal{S}^0_k \cap \mathcal{S}_k}$. To make this practically feasible, we assume that the majority of the data points in a cluster are correctly assigned. Theoretically, at least half of the data points need to be correctly assigned to guarantee an improvement. However, our experiments suggest that for computational reasons, in practice it is desirable to have at least 80--90\% of correct assignments, though the exact behaviour of the method will further depend on the actual geometry of the data distribution.  

If the index set $\mathcal{S}^0_k \cap \mathcal{S}_k$ were known, this would give the Oracle the knowledge of the best approximate subspace given by $\mathsf{span}(\bs{X}_{\mathcal{S}^0_k \cap \mathcal{S}_k})$. A better estimate of the true subspace in the absence of other information cannot be obtained. Therefore this estimate defines the best achievable reference, which is useful for the evaluation of the proposed method.   

\begin{figure}
 	\centering
	\includegraphics[trim={0.5cm 0 1cm 0},clip,width=.99\linewidth]{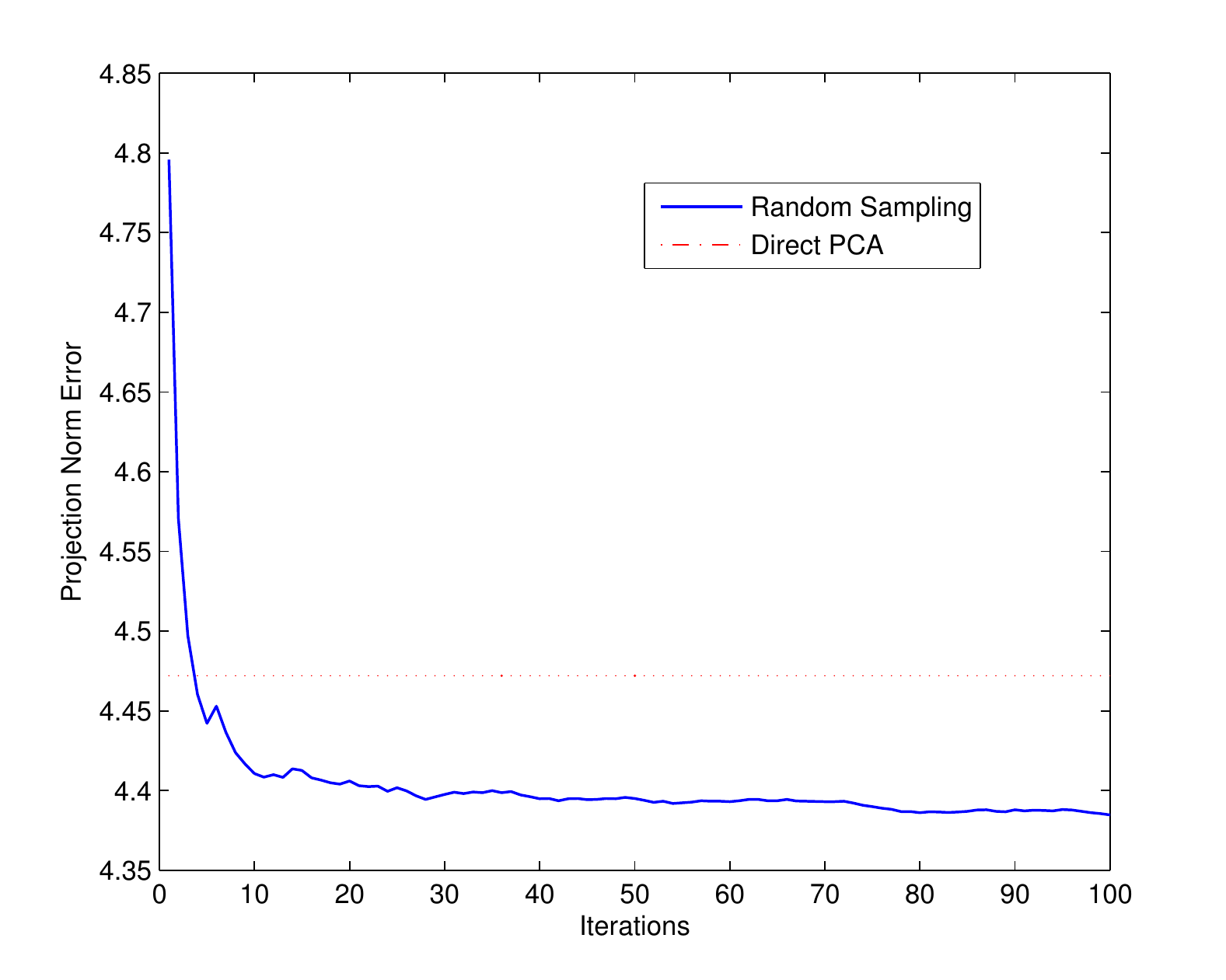}
	\caption{\label{FIG_SAMPLING} An illustrative example showing the convergence behaviour of the projection matrix computed from a random sampling scheme on synthetic data, wherein the majority of the samples follow a true subspace model, except for a small fraction $\alpha$. The y-axis shows the Frobenius norm of the error with respect to that obtained by the Oracle knowledge.  }
\end{figure}

In this work, we do not directly estimate the subspace spanned by $\mathcal{S}^0_k \cap \mathcal{S}_k$. Instead, we compute the approximate projection on $\mathsf{span}(\bs{X}_{\mathcal{S}^0_k \cap \mathcal{S}_k})$ through the process of random sampling. To motivate this idea using an illustrative example, consider the following synthetic problem which demonstrates the mechanism behind the stable subspace learning method which we will explain in detail thereafter. Here, we generate a synthetic cluster data of $N=100$ samples in $\mathbb{R}^d, d=100$, wherein $1-\alpha$ fraction of the samples belong to the true subspace of the dimension $10$. The remaining $\alpha$ fraction contains outliers which are uniformly distributed in $\mathbb{R}^d$. We also left-multiply the data by a random unitary matrix to ensure the final data is not trivial. The goal is to learn the true subspace in the presence of outliers. The subspace is learnt by computing its projection. We consider two methods. The first method is in the form of principal component analysis (PCA) that extracts the principal subspace on the whole cluster data using a principal energy fraction of $\rho=1-\alpha$. This yields the projection $\bs{P}_p = \bs{U}_p\bs{U}_p^T$ where $\bs{U}_p$ is the left singular sub-matrix corresponding to the principal components. The second method is a variation of PCA: we randomly select a fraction $\rho$ of the cluster data, extract the corresponding projections onto the principal subspace of that subset, and then compute the average value of the projection matrices for each iteration, which we denote as $\bs{P}_s$. We then compare the computed projection matrices with that obtained by the Oracle's knowledge of the relevant samples within the clusters, which is denoted as $\bs{P}_t$. Figure~\ref{FIG_SAMPLING} shows how the Frobenius norm error $\|\bs{P}_s-\bs{P}_t \|_F^2$ reduces when the number of iteration increases. It also shows the error of direct PCA, which is $\|\bs{P}_s - \bs{P}_p \|_F^2$. The figure clearly shows that the random sampling process generally yields an improved estimate of the projection matrix as the number of iterations increases, and that the stable projection matrix achieves a smaller relative error than does conventional PCA. 

Formalizing the procedure from the example given above, the proposed method can be summarized by the following sequence of steps: 

\noindent\textbf{Step 1: Obtaining stable principal subspaces from the preliminary clusters}
\begin{itemize}
 \item For each cluster $C_k$ repeat the following for iteration  $i = 1, \ldots, maxIter $
	\begin{itemize}
		\item Randomly select a subset $\bs{X}^k_i$ of size $\rho N_k$ from $\bs{X}^k$, where $\rho$ is a number close to 1, which designates the fraction of the correctly clustered samples;
		\item Obtain the principal eigenmatrix $\bs{U}^k_{P}$ for an energy fraction $\rho$ over this random subset;
			\begin{itemize}
				\item Perform singular value decomposition:
						\begin{eqnarray}
							\bs{X}^k_i & = & \bs{U}\bm{\Sigma}\bs{V}^T,\\
							 \bm{\Sigma}& = & \diag(\sigma_1,\sigma_2,\ldots,),\\
							 	&& \sigma_1 \geq \sigma_2 \ldots.
						\end{eqnarray} 
				\item Find minimum $P$ such that:
						\begin{eqnarray}
							\frac{\sum_{l=1}^{P}\sigma_l}{\sum_{l}\sigma_l} \geq \rho.							
						\end{eqnarray}
				\item Select $\bs{U}^k_{P}$ as the columns of $\bs{U}$ corresponding to $\sigma_1,\ldots,\sigma_P$ 			
									
			\end{itemize}
		\item Construct the residual projection as $\bs{P}_i^k=\bs{I}-\bs{U}^k_{p}{\bs{U}^k_{p}}^T$;	
	\end{itemize}
 \item For each cluster $k$, compute the stable projection onto the residual subspace:
		\[ \bs{P}_R^k= \frac{1}{maxIter} \sum_{i=1}^{maxIter}\bs{P}_i^k. \]
			
\end{itemize}

\noindent\textbf{Step 2: Dominant nearest subspace clustering}
\begin{itemize}
 \item For each data point $\bs{x}_n$, compute the residual vector for each subspace $k$:
	\[ \bs{r}^k_n = \bs{P}_R^k \bs{x}_n.\]
 \item Compute the residual score of each data point to each subspace as the $\ell_p$-norm of the corresponding residual vector:
	\[ e^k_n = \| \bs{r}^k_n\|_p.\]
 \item Denote the set $\mathcal{E} = \{ e^k_n: k=1, 2, \ldots\}$ Without loss of generality, suppose that $e^1_n$ is the residual score for the subspace computed from the current clusters and $e^2_n, e^3_n \ldots$ are those for the subspaces computed from the other clusters. In dominant nearest subspace clustering, we only consider the re-assignment of the current data point if the best residual error from other clusters is considerably less than that of the currently-assigned cluster:
		\[ \min_{k\geq 2} e^k_n \leq \eta e^1_n, \] 
where $0<\eta<1$ is a small number that quantifies the notion of ``significantly smaller''. If this condition is satisfied, we re-assign the data point to the cluster having minimum error:
	\[ C(\bs{x}_n) = \arg\min_{k \geq 2} e^k_n. \]
\end{itemize}

\begin{figure}[t!]
 	\centering
	\includegraphics[trim={1cm 2cm 1cm 1.8cm},clip,width=.99\linewidth]{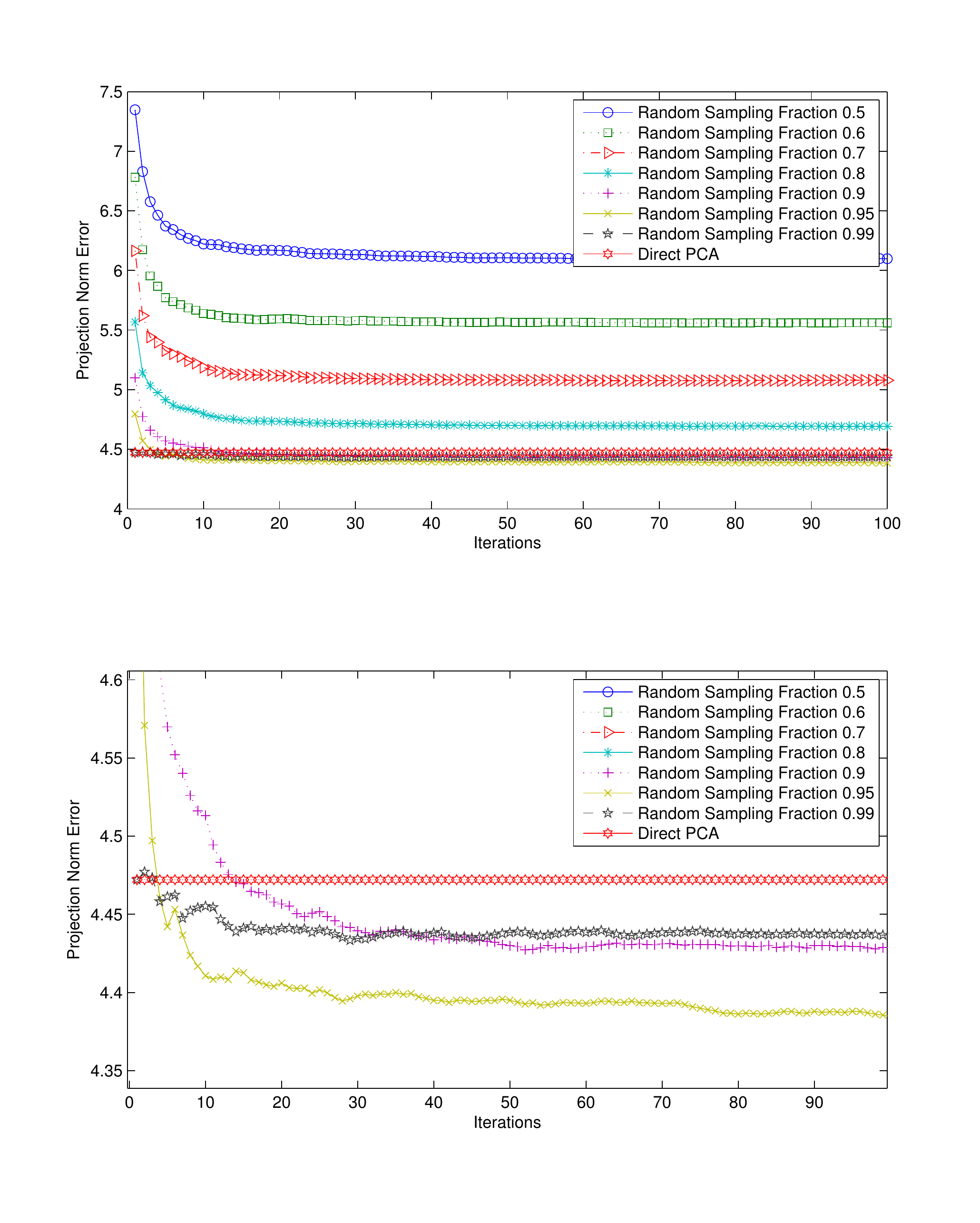}
	\caption{\label{FIG_FRACTION} An illustrative example on synthetic data showing the effect of varying the size of the randomly sampled subsets as a fraction of the original cluster. Here, the true fraction of samples correctly clustered is $\rho=0.95$. It is clear that sampling at lower sizes results in a worse steady-state error. Sampling at the correct value of $\rho$ achieves optimal performance. The bottom subplot shows a magnification of the top subplot for sampling fractions around the optimal value.}
\end{figure}

%\noindent\textit{Remarks}
\paragraph{\it Remarks}
\begin{itemize}
 \item In the algorithm summarized above, $maxIter$ is the number of repetitions in Step~1. Alternatively, one may also check if the projection matrix $\bs{P}_R^k$ converges to some stable value so as to terminate the iteration early. 
 \item One key parameter of the algorithm is the energy fraction of the principal subspace $\rho$ that controls the complexity of the underlying predictive model. A high value of $\rho$ likely results in subspace over-fitting, whilst a low value generally leads to a noisier prediction. Our experiments suggest that $\rho=0.9$ achieves good results across many data types tested. Of course, it is desirable to extend to the case where $\rho$ may vary between clusters in order to provide more comparable fitting errors between them. This task is left for future work. 
 \item Another parameter of the algorithm is the size of the subset. Here, we choose it to be exactly the fraction of the samples correctly clustered i.e.\ $\rho$. In the literature on stability selection, it is often the case that a randomly selected half of a cluster is being sampled at a time. However, numerous experiments that we conducted suggest that $\rho$ is the optimal choice for the size of the randomly sampled subsets. Figure~\ref{FIG_FRACTION} supports this observation.
 \item In the second step, the $\ell_p$-norm is used to compute the deviations of the data points from the subspaces. Here, we suggest $p=1.5$ for a good balance between dense and sparse errors, which is observed to provide an overall satisfactory performance in many cases.
 \item The proposed method is only useful if the preliminary subspace clustering result is sufficiently accurate in the sense that the size of each cluster found is about the expected size so that the purity in each cluster exceeds 50\%. This allows the principal subspace from each cluster to be extracted stably and reliably.
 \item In the proposed method, we introduced a new concept which we termed \emph{dominant nearest subspace clustering}, particularly designed as a post-processing technique. The idea is that a re-assignment of a data point to a new cluster is necessary if that data point much better fits another stable subspace. This is critical as it guards against noise and unavoidable errors when the stable subspaces are extracted. This process is governed by the parameter $\eta$. Clearly, the smaller the value of $\eta$, the more conservative the scheme is. When $\eta=1$, the process reduces to the conventional nearest subspace classification. A large value of $\eta$ may correct more data points, but potentially introduces disturbance to correct cluster assignments achieved in preliminary clustering. Similarly, a small value of $\eta$ may achieve less correction, but is safer as it minimizes the aforementioned disturbance. In this work, we used $\eta=0.5$.
\end{itemize}

\section{Empirical Evaluation}
\label{SEC_EXP}
In this section we demonstrate experimentally how the proposed method improves preliminary clustering results by sparse subspace clustering (SSC) \cite{ElhaVida2013}. We use the original code provided by the authors of SSC and set the parameters using the default values in these implementations. We consider three popular subspace clustering datasets: the Johns Hopkins 155 motion segmentation dataset\footnote{\scriptsize{\url{http://www.vision.jhu.edu/data/hopkins155/}}}, the CMU pose, illumination, and expression (PIE) dataset\footnote{\scriptsize{\url{http://vasc.ri.cmu.edu/idb/html/face/}}}, and the extended Yale B face dataset\footnote{\scriptsize{\url{http://vision.ucsd.edu/~leekc/ExtYaleDatabase/ExtYaleB.html}}}. The two face datasets were originally collected for evaluation of face recognition algorithms (for the best reported recognition performance on this data set see \cite{Aran2012e}) but have since also been widely adopted in the literature on subspace clustering. Their suitability within this context stems from the finding that under the assumption of the Lambertian reflectance model the appearance of a face in a fixed pose is constrained within the corresponding image space to a 9-dimensional subspace~\cite{GeorBelhKrie2001} (this observation is used extensively in numerous successful manifold based methods such as \cite{Aran2014b,Aran2014a,AranHammCipo2010}).\\

To demonstrate the usefulness of the proposed method, we evaluate clustering results using classification errors (as done in the original work on SSC), normalized mutual information (which is a popular performance metric in the general clustering literature), and the number of corrections and false re-assignments of data points. This range of performance measures provides a comprehensive picture of the behaviour of the proposed method.

\subsection{Experiment 1: Motion Segmentation} 

We first consider the motion segmentation problem. The Hopkins 155 dataset consists of 2- and 3-motion sequences, each being a collection of the $(x,y)$ coordinates of tracked points on moving objects captured by a (possibly moving) camera. The dataset has been used as a standard subspace clustering benchmark, including in the original work on SSC.

\begin{figure}[t!]
 	\centering
	\includegraphics[trim={1.5cm 2cm 1cm 1cm},clip,width=\linewidth]{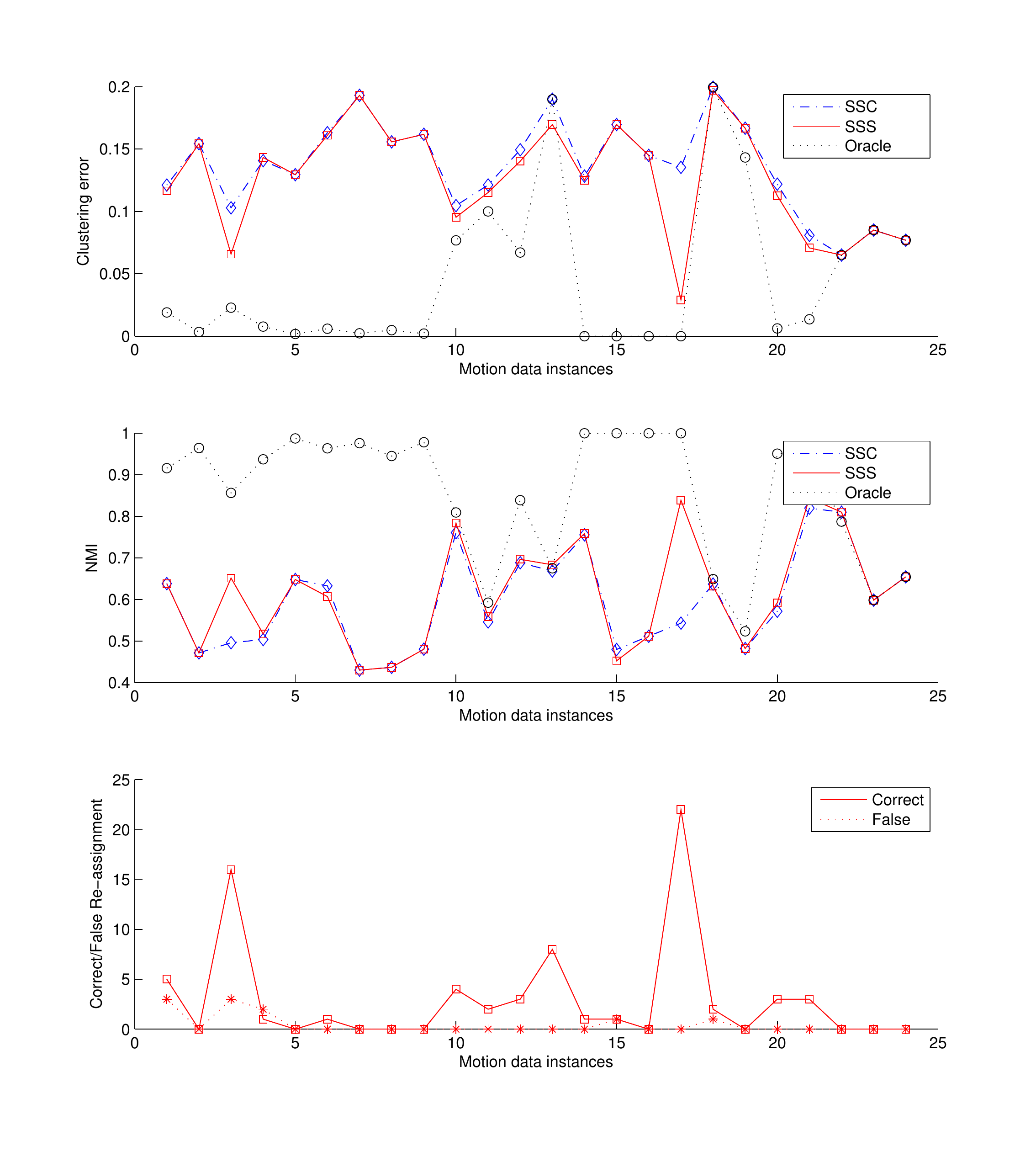}
	\caption{\label{FIG_MS} Summary of our re-clustering results on the Hopkins 155 motion segmentation dataset.}
\end{figure}

We adopt the default setting for SSC: the regularization parameter is set to 20, and no dimensionality reduction or affine constraints are used. With this setting, we obtain the preliminary clustering result by SSC on all sequences and then use the proposed method to refine it. We show the re-clustering result for sequences whose primary clustering error is between 0.05 and 0.2 in Figure~\ref{FIG_MS}. Here, we plot the before and after values for clustering errors in the top subplot, NMI in the middle subplot, and the number of correct versus false re-assignments in the bottom subplot. Performance improvement is achieved when the clustering error is reduced and NMI is enhanced. This is observed markedly in at least 8 sequences, where the proposed method makes significantly more corrections than introducing re-assignment errors. The best example is when the proposed method corrects 23 data samples whilst introducing no re-assignment error at all. In this case, the average clustering error reduces from more than 10\% to less than 5\%. For other motion sequences, the conservative strategy seems to be in effect as the proposed method makes few changes to the preliminary clustering. Figure~\ref{FIG_MS} also shows the performance of the similar method but with the subspaces obtained from the Oracle's knowledge of the true data samples within a found cluster. This establishes the maximum achievable performance that the proposed method can achieve. As can be seen, there are few cases when the proposed method is quite close to the bound. However, a majority of cases indicate that there is still a significant gap, which clearly motivates future work that can better estimate the subspaces.

\begin{table*}
	\centering
	\renewcommand{\arraystretch}{1.4}
	\caption{\label{TAB_YALEB} Face clustering results on the Yale B dataset.}
	\begin{tabular}{|c|c|c|c|c|c|c|c|c|c|c|} \hline
		\multicolumn{1}{|c|}{ } &  \multicolumn{3}{c|}{Error} & \multicolumn{3}{c|}{NMI} & \multicolumn{4}{c|}{Re-Assignment}  \\ \hline\hline 
		        & SSC & SSS & OSS   & SSC & SSS & OSS  &  \multicolumn{2}{c|}{SSS} & \multicolumn{2}{|c|}{OSS}  \\ \hline 
		Classes	&     &     &       &     &     &      &     Correct & False  & Correct & False  \\ \hline
		2       & 0.0625 & 0.0469 & 0.0391& 0.6653 &0.7270 & 0.7723 & 2 & 0 & 3 & 0 \\
		3       & 0.0833 & 0.0833 & 0.0729& 0.7671 &0.7671 & 0.7831 & 0 & 0 & 2 & 0 \\
		5	& 0.0656 & 0.0563 & 0.0531& 0.8291 &0.8501 & 0.8547 & 3 & 0 & 4 & 0 \\         
		8 	& 0.0566 & 0.0488 & 0.0488& 0.8866 &0.9059 & 0.9036 & 4 & 0 & 4 & 0 \\ \hline
	\end{tabular} 	
\end{table*}

\begin{table*}
	\centering
	\renewcommand{\arraystretch}{1.4}
	\caption{\label{TAB_PIE} Face clustering results on the PIE dataset.}
	\begin{tabular}{|c|c|c|c|c|c|c|c|c|c|c|}  \hline
		\multicolumn{1}{|c|}{ } &  \multicolumn{3}{c|}{Error} & \multicolumn{3}{c|}{NMI} & \multicolumn{4}{c|}{Re-Assignment}  \\ \hline\hline 
		        & SSC & SSS & OSS   & SSC & SSS & OSS  &  \multicolumn{2}{c|}{SSS} & \multicolumn{2}{|c|}{OSS}  \\ \hline 
		Classes	&     &     &       &     &     &      &     Correct & False  & Correct & False  \\ \hline
		2       & 0.0250 & 0.0.0250 & 0.0000& 0.8858 &0.8858 & 1.0000 & 0 & 0 & 1 & 0 \\
		3       & 0.0781 & 0.0677 & 0.0417& 0.7524 &0.7859 & 0.8357 & 2 & 0 & 7 & 0 \\
		5	& 0.0938 & 0.0875 & 0.0844& 0.7851 &0.7988 & 0.8084 & 2 & 0 & 3 & 0 \\         
		8 	& 0.0688 & 0.0688 & 0.0625& 0.9486 &0.9486 & 0.9501 & 0 & 0 & 1 & 0 \\ \hline
	\end{tabular} 	
\end{table*}

\subsection{Experiments 2 and 3: Face Clustering}

We next consider the face clustering problem. Unlike the motion segmentation dataset which is limited to only 3 motions, the number of different persons in the Yale B and PIE face datasets is larger. In this work, we consider 2, 3, 5, and 8 classes for the face clustering problem. For each run, we randomly select the specified number of persons from the face datasets and obtain clustering results using SSC. The ideal clustering result is when all images of the same person fall into a cluster. The parameters for SSC are the same as in the previous experiment. \\

Tables~\ref{TAB_YALEB} and~\ref{TAB_PIE} show the re-clustering results obtained by the proposed stable subspace (SSS) and ordinary subspace (OSS) approaches on these two face datasets. In all cases, once again we notice the conservative strategy is effective in avoiding re-assignment errors, whilst the stable subspace selection mechanism helps identify and correct cluster outliers. There are only three cases (3-class in Yale B and 2-class and 8-class in PIE) where the proposed SSS does not make any changes to the preliminary clustering results. Otherwise, it provides further improvement even when the number of the classes is large (e.g.\ 8 persons in Yale B). Here, a number of interesting observations can be made. Firstly, the number of corrections made by SSS is close or at the same order of magnitude as that by OSS, i.e.\ with the Oracle knowledge of the subspaces. Secondly, the ability of SSS to make corrections depends on a number of factors, and thus having either a small number of classes or good preliminary clustering does not automatically guarantee further refinement. For example, SSS still makes 2 corrections for an initial 6.25\% clustering error in the case of Yale B with 2 classes. However, with two classes and the initial clustering error of 2.5\%, it does not make a further correction on PIE. Likewise, for both data sets with 8 classes at similar initial clustering errors, SSS is able to achieve maximum correction on Yale B, whilst it cannot improve any further on PIE. We note that it is possible to increase the number of correct re-assignments for SSS, by increasing the parameter $\eta$. However, we still suggest the conservative setting to ensure the effectiveness of this post-processing step.

\section{Summary and conclusions}
\label{SEC_CONC}
We proposed a novel method for refining and improving sparse subspace clustering. The key idea underlying the proposed method is to learn a stable subspace from each cluster by random sampling and then re-evaluating how well each data point fits the subspaces model by computing the residual error corresponding to each of the initial subspaces. Dominant nearest subspace classification, a conservative strategy, is then used to decide whether or not a data point should be assigned to a more suitable cluster. Experiments on widely used data sets in subspace clustering show that the proposed method is indeed highly successful in improving the results obtained by traditional sparse subspace clustering. Our future work will address how to obtain a better subspace extraction to bring the performance closer to that of the Oracle.

\balance

\bibliographystyle{ieeetran}
\bibliography{../../../my_bibliography}

% that's all folks
\end{document}